# A toolkit for a Generative Lexicon

**Patrick HENRY**
CNRS - LaBRI - INRIA - UMR5800
Université de BORDEAUX
351, Cours de la Libération
TALENCE, FRANCE, 33405
henry@labri.fr

**Christian BASSAC**
INRIA - LaBRI - CNRS - UMR 5263
Université de BORDEAUX
351, Cours de la Libération
TALENCE, FRANCE, 33405
Christian.Bassac@inria.fr

**Abstract**

In this paper we describe the conception of a software toolkit designed for the construction, maintenance and collaborative use of a Generative Lexicon. In order to ease its portability and spreading use, this tool was built with free and open source products. We eventually tested the toolkit and showed it filters the adequate form of anaphoric reference to the modifier in endocentric compounds.

## 1 Introduction

Over the past ten years or so, the Generative Lexicon (from now on GL: Pustejovsky (1995)), has been used, among other things, as a new tool in lexical semantics. The representation language it provides has been used to investigate various linguistic phenomena, in various languages (see for instance *Proceedings of The first/second/third International Workshop on Generative Approaches to the Lexicon 2001/2003/2005*) and an increasingly unified and powerful formalism emerged (Asher and Pustejovsky forthcoming). A GL can also be considered as a promising theory which can be applied to build NLP tools (Gupta and Aha 2003), for instance with a view to validating linguistic rules on large corpora. The aim of this paper is to present an example of such a tool: basically it is a toolkit which will be used to manipulate a GL for further applications. Our presentation will go into some detail about the data structures and the software that we chose in order to build, maintain and use a GL collaboratively.

## 2 The software: objectives and constraints

First we set ourselves three constraints:

- We think that the task of building resources should be mutualized, due to the amount of work that must be put in. Although works on the automatic acquiring of qualia structure do exist, (see for instance (Sébillot 2002), we think that the final adjustments must be hand-made. So, right from the outset we deliberately chose a system that allows collaborative work.
- It also seems to us that this toolkit should be available and used on the usual platforms that cover most current standards, WINDOWS, UNIX/LINUX, and MAC/OS.
- The kit must also be as user-friendly as possible and the building of a lexicon must be both intuitive and interactive. We also wish that the user can visualize the precise contents of the resources, consequently we think that building and maintenance should be carried out with graphic interfaces.

## 3 The tools used

Taking into account the constraints previously described, we chose the following tools:

LDAP[http://www.opendap.org/]: This protocol directly derived from X500 was initially conceived to build electronic directories. It can be observed that there are similarities between a directory and a lexicon, for instance both allow access to information from an entry. This protocol seemed to be the right thing to us as it has the following functionalities:

- It can be mutualized
- Data structures can be parametrized
- Its consultation is optimized for speed access
- It can be used in network
- Toolkits are ready to use and free
- Its use is cross-platform

PYTHON [http://www.python.org/]: The choice of the language is of paramount importance: in this type of application the

specifications keep evolving as work goes on, and consequently an easy-handling device is needed, which also eases further work on the software. So we decided to use an interpreter. Among the functionalities we need, object-oriented programming is crucial as it allows the modification of some data structures without having to go back to the code already written. Among the functionalities that we deem relevant with PYTHON, the following are of the utmost interest:

- The interpreter is among the most efficient to-date (semi-compiled)
- Object-oriented programming
- The WXWINDOW graphic toolkit is available (WXPYTHON).
- The LDAP toolkit is available
- The language handles lists
- The language has lambda-expressions

JXPLORER [http://www.jxplorer.org/]: It is a simple LDAP browser-editor written in JAVA that can be used on any platform. This is the tool we used in the construction and maintenance of the lexicon. Figure 1 below shows an example of the editing of a lexical entry: the item is chosen in the left hand column, and the details of the entry are visualized in the central part, the nomenclature being inspired by the notational conventions of GL. The terminology somewhat differs from classical GL notation on three points: first the various qualia roles are not grouped together, second the key-words TRIGGER and RESULTAT (result) are used respectively for the AGENTIVE and FORMAL roles embedded in the TELIC found in various GL works, and third some variables such as OBJECTCLASS are inherited from the LDAP data structure and are not directly connected with GL. A partial view of the construction of a lexical entry is given below in figure 1.

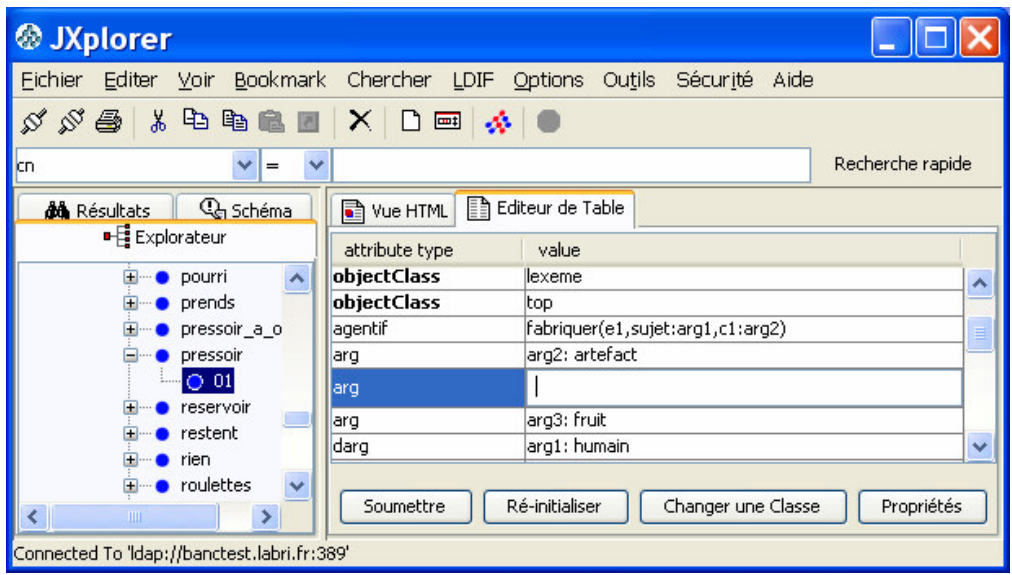

Fig 1: Partial view of the construction of a lexicon entry.

It is fairly easy to add a value by a right click on the type of the value to add. We chose to show the example of the construction and editing of this particular item because it is part of the lexicon in the theoretical study by Bassac and Bouillon (2005, 27) we later used to test our toolkit[1]. The final view of a lexicon entry is given in figure 2 below.

---

[1] Bassac and Bouillon (2005) examined the data of French, English and Turkish but the validity tests here are carried out on French data only. The entry chosen as an example here is the word *pressoir* [press]$_N$ (like in "cider press", see 5 below).

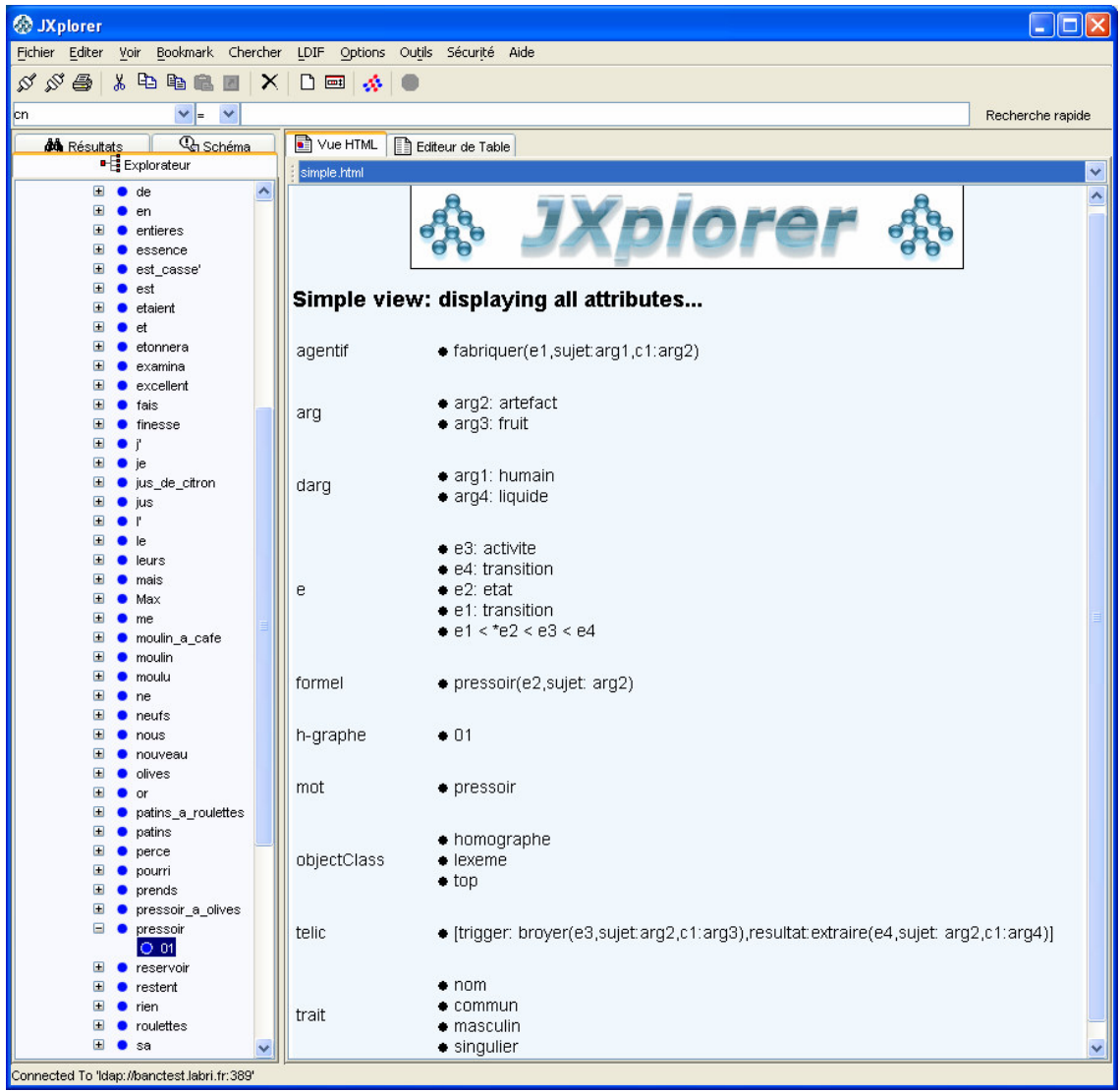

Figure2: View of a lexicon entry with JXPLORER

Although, the tests were carried out on a limited number of lexicon entries, the extension to large scale applications should be easy as LDAP has already been optimized for quick access to data.

**4      Functionalities available**

The programs allow the following operations :

- On the LDAP server (DSA): Centralization and acceptance of requests for consultation or modification of the lexicon. The software can be configured so as to include access-control.
- The JXPLORER (DUA) client allows connexion to the lexicon in order to create, modify or search an entry in the lexicon.
- The entree.py (DUA) client is a python module (library) which allows the following operations :

    o   Connexion of the program to the lexicon, possibly with authentication
    o   Saving and restoring the lexicon in LDIF or XML format
    o   Searching a word in the lexicon
    o   Retrieval of the features of an entry.
    o   Retrieval of a feature value
    o   Pretty printing of a lexicon entry

The architecture of the whole system is summed up and schematized in figure 3 below:

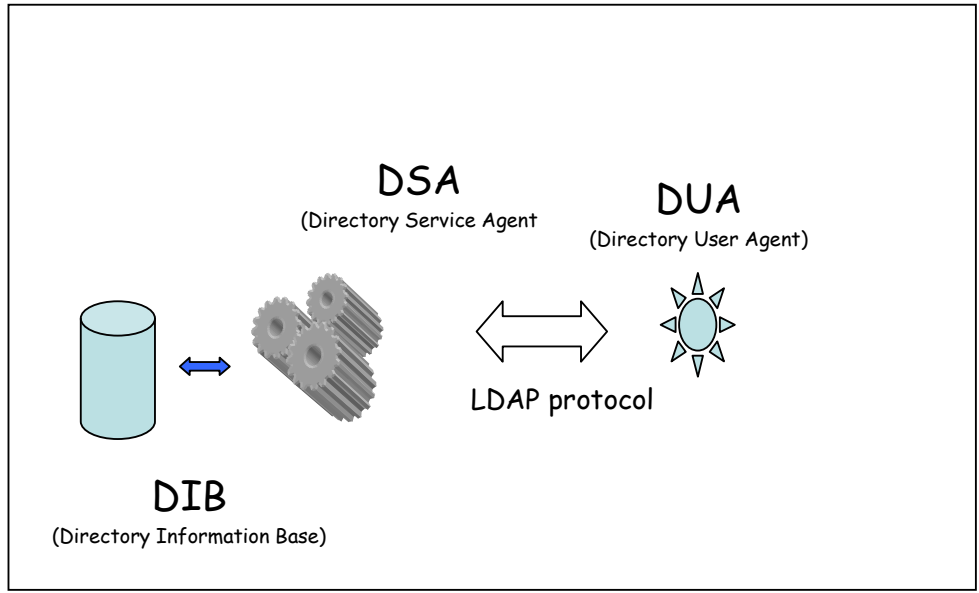

Figure 3: The client-server architecture of the lexicon

## 5  A comparison with other existing toolkits for lexicon input.

Other similar projects do exist, whose aims are the constructions of lexicons. One of these projects is the ISLE project (Bel, Villegas, Marimon (2003)) that terminated in 2003. It is an interesting and relevant project in so far as it led to the development of tools used in the implementation of lexicons, which is our objective too. Yet, the final goals of the ISLE project were slightly different: the lexicons were multilingual, and the information content was not uniform and depended on the entry, mainly in order to perform disambiguation tasks. The whole system is a kind of meta-entry in which lexical information is not distributed according to the principles of a GL, but is layered with a view to easing the process of automatic translation. Our toolkit, here presented, was specially designed for the development of applications that conform fairly strictly to the principles of GL. More specifically, we concentrated on these aspects of lexical semantics that allow coherence checking operations and validation of linguistic hypotheses. The implementation we chose also departed from the ISLE project as it was built on data bases which can be accessed via PERL CGI scripts and a WEB browser. Our approach is more orientated towards the implementation of NLP applications such as coherence checking or generation. This is why our toolkit heavily relies on the choice of LDAP and PYTHON.

## 6  An application

We chose to test our scale model on the paper by Bassac and Bouillon (2005): the reasons for our choice are that first in this paper the authors offer an explanatory account of an interesting syntactic phenomenon, and second the range of vocabulary items used is both varied and its number can be easily handled (seventy or so).

The authors show that anaphoric reference to the modifier of a compound is qualia-driven and depends both on the type of the relation R that holds between the modifier $N_1$ and the head $N_2$, and the role it is encoded in:

- If R is a `contain` relation, then anaphoric reference to $N_1$ is possible via a definite determiner NP (DeFDetNP) but not via a possessive determiner NP (PossDetNP) nor via a demonstrative determiner NP (DemDetNP).
- If R is a `part_of` relation, anaphoric reference to $N_1$ is possible via a definite determiner NP or a possessive determiner NP.
- If $N_1$ saturates the second argument of a predicate encoded in the agentive ("trigger" in our lexicon) of the telic, then anaphoric reference to $N_1$ is possible only via a definite determinant NP.
- If $N_1$ saturates the first argument of a predicate encoded in the formal ("result" in our lexicon) of the telic, then anaphoric reference to $N_1$ is possible only via a definite determiner NP or a possessive determiner NP.
- If R is encoded in the agentive, then anaphoric reference to $N_1$ is possible only via a definite determiner NP.

This is summed up in figure 4 below:

| Relation | Tel=state | Part_of | Tel,Agent | Tel,Form | Agentive |
|---|---|---|---|---|---|
| Example | *verre à vin* (wine glass) | *patin à roulettes* (roller skates) | *pressoir à olives* (olive press) | *pressoir à cidre* (cider press) | *jus de citron* (lemon juce) |
| Anaphoric Relation | DeFDetNP *PossDetNP *DemDetNP | DeFDetNP PossDetNP *DemDetNP | DeFDetNP *PossDetNP *DemDetNP | DeFDetNP PossDetNP *DemDetNP | DeFDetNP *PossDetNP *DemDetNP |

Figure 4: head/modifier relations and anaphoric reference to the modifier

The authors show that this accounts for the following facts:

(1) Le verre a vin est cassé, * son vin coule.
  The wine glass is broken, *its wine is leaking out of the crack.
(2) Max examina ses patins à roulettes : leurs roulettes étaient tout usées.
  Max took a look at his roller skates: Their rollers were worn-out.
(3) Ce pressoir est défectueux, *ses olives restent entières.
  This olive press does not work properly, *its olives are not properly ground.
(4) Nous utilisons un nouveau pressoir, son cidre est excellent.
  We use a new cider press, its cider is great.

In our implementation we also built and used a partial type hierarchy in order to control the unification of the types of variables that appear in the qualia structure.

The facts exemplified in (1) to (4) above are validated by our system as shown below for example (3) and (4):

```
sh-2.05b$ ./demo3
+ python te2a.py -b pressoir olives 'Ce %s est défectueux; %s %s restent entières.'

Ce programme est une maquette pour valider
une implémentation du lexique génératif.

Le jeu d'essai ainsi que les concepts utilisés proviennent de l'article
"Qualia structure and anaphoric reference in compounds"
GL2005, May 19 - 21, 2005

Il accepte deux mots et un modele en entrée, puis génère
les validités des reprises anaphoriques.

  Ce pressoir est défectueux; les olives restent entières.
* Ce pressoir est défectueux; ses olives restent entières.
* Ce pressoir est défectueux; ces olives restent entières.

+ python te2a.py -b pressoir cidre 'Nous utilisons un nouveau %s, %s %s est excellent.'

Ce programme est une maquette pour valider
une implémentation du lexique génératif.

Le jeu d'essai ainsi que les concepts utilisés proviennent de l'article
"Qualia structure and anaphoric reference in compounds"
GL2005, May 19 - 21, 2005

Il accepte deux mots et un modele en entrée, puis génère
les validités des reprises anaphoriques.

  Nous utilisons un nouveau pressoir, le cidre est excellent.
  Nous utilisons un nouveau pressoir, son cidre est excellent.
* Nous utilisons un nouveau pressoir, ce cidre est excellent.

sh-2.05b$
```

Figure 5: Execution on a computer

## 7. Assessment and future work

As we have shown, this software is operational and has already been used in the construction of an NLP application. It can be readily used in the building and maintenance of a GL. One of the limits of the system is the size of the lexicon and the obligation to encode the lexicon information manually. Consequently the next steps will lead us to code wider scale mutualizable resources. For instance, we plan to use corpora or systems such as WordNet and its ontology to automatically acquire the relevant information necessary to build qualia structures (see for instance Sébillot 2002). This task is fairly difficult, but it can be eased by the use of morphological indications contained in a corpus (see Namer, Bouillon, Jacquey, this volume) or by syntactic clues: for instance a sequence $N_1$ *de* $N_2$ in French in which $N_1$ is a container and $N_2$ is a liquid is an example of dot introduction.

The ergonomy of the system can be improved so as to make it more user-friendly. In the near future JXplorer which was used in the first steps of the development of the system should be given up for a tool which will be easier to use, so as to show only the relevant information. For large scale use we have planned to develop interfaces that allow the definition of rules rather than algorithms as well as checking procedures for the coherence of lexical entries.